\documentclass{article}

\PassOptionsToPackage{numbers, sort&compress}{natbib}
\usepackage[preprint]{neurips_2023}

\usepackage[utf8]{inputenc} 
\usepackage[T1]{fontenc}    
\usepackage{url}            
\usepackage{booktabs}       
\usepackage{amsfonts}       
\usepackage{nicefrac}       
\usepackage{microtype}      

\usepackage{graphicx}
\usepackage{subfigure}
\usepackage{amsmath}
\usepackage{amssymb}
\usepackage{booktabs}
\usepackage{bbm}
\usepackage[dvipsnames,svgnames,x11names,table]{xcolor} 
\usepackage{pifont}
\usepackage{bbding}

\usepackage{wrapfig}
\usepackage{multirow}
\usepackage{transparent}
\usepackage{tikz}

\definecolor{pretty-blue}{RGB}{0, 113, 188}
\definecolor{brown}{RGB}{201, 104, 71}

\def\eg{{\it{e.g.}}}

\def\ie{{\it{i.e.}}}
\def\etc{{\it{etc}}}

\usepackage[colorlinks=True,
            linkcolor=blue,
            anchorcolor=blue,  
            pagebackref,
            urlcolor=blue,
            citecolor=teal,
            ]{hyperref}

\title{\ \ ChatSpot: Bootstrapping Multimodal LLMs via Precise Referring Instruction Tuning }

\author{
Liang Zhao$^{1}$\thanks{Equal contribution}~,
En Yu$^{2*}$, Zheng Ge$^{1}$\thanks{Project leader}, \bf{Jinrong Yang}$^{2}$, Haoran Wei$^{1}$,\\
\bf{Hongyu Zhou$^{1}$}, \bf{Jianjian Sun}$^{1}$, Yuang Peng$^{3}$, \bf{Runpei Dong}$^{4}$, \textbf{Chunrui Han}$^{1}$, \textbf{Xiangyu Zhang}$^{1}$ \\\\
$^{1}$MEGVII Technology, $^{2}$Huazhong University of Science and Technology  \\
$^3$Tsinghua University, $^4$Xian Jiaotong University  \\\\
{Demo: \url{https://chatspot.streamlit.app}}
}

\begin{document}

\maketitle

\begin{tikzpicture}[remember picture,overlay,shift={(current page.north west)}]
\node[anchor=north west, xshift=3.0cm, yshift=-3.0cm]{\scalebox{-1}[1]{\includegraphics[width=1.5cm]{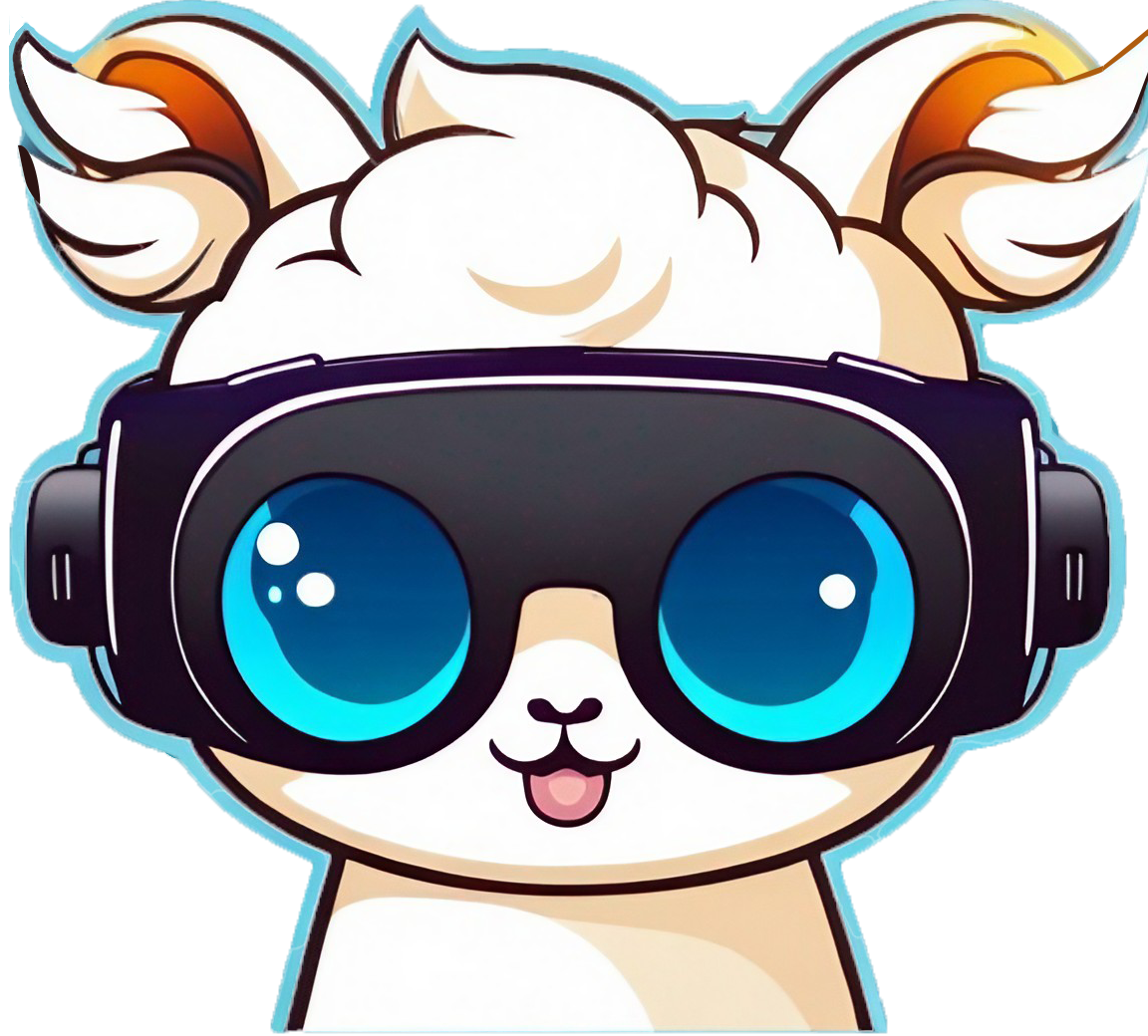}}};
\end{tikzpicture}



\begin{abstract}
Human-AI interactivity is a critical aspect that reflects the usability of multimodal large language models (MLLMs). However, existing end-to-end MLLMs only allow users to interact with them through language instructions, leading to the limitation of the interactive accuracy and efficiency. In this study, we present \textit{precise referring instructions} that utilize diverse reference representations such as points and boxes as referring prompts to refer to the special region. This enables MLLMs to focus on the region of interest and achieve finer-grained interaction. Based on precise referring instruction, we propose ChatSpot, a unified end-to-end multimodal large language model that supports diverse forms of interactivity including mouse clicks, drag-and-drop, and drawing boxes, which provides a more flexible and seamless interactive experience. We also construct a multi-grained vision-language instruction-following dataset based on existing datasets and GPT-4 generating. Furthermore, we design a series of evaluation tasks to assess the effectiveness of region recognition and interaction. Experimental results showcase ChatSpot's promising performance.
\end{abstract}

\section{Introduction}
\label{intro}

Recent advances in large language models (LLMs) exemplified by GPT-3~\cite{GPT3} and LLaMA~\cite{llama} have demonstrated significant potential in the domain of zero-shot learning and logical reasoning. By aligning pre-trained LLMs to follow human language instructions through Reinforcement Learning with Human Feedback (RLHF)~\cite{christiano2017deep}, InstructGPT~\cite{InstructGPT} and ChatGPT~\cite{ChatGPT} have showcased powerful capabilities for human-AI interaction, leading to a new paradigm shift towards the realization of artificial general intelligence (AGI).

Inspired by the remarkable success of GPT series~\cite{GPT3, ChatGPT, GPT4}, researchers attempt to incorporate more modalities into LLMs for multimodal human-AI interaction, with vision-language interaction being an important topic of focus. In order to incorporate visual modality into LLM, significant processes have been made to bridge the gap between LLMs and vision foundation models. There are two mainstream paradigms for building multimodal large language models (MLLMs). One is \textit{plugin-based} MLLM~\cite{VisualChatGPT, MMREACT, Hugginggpt} that utilizes off-the-shell LLMs~\cite{ChatGPT, llama, vicuna} as central controllers to schedule different visual expert models as plugins. In this way, the users can interact with LLMs to achieve diverse visual functions. Another paradigm is \textit{end-to-end} MLLM~\cite{Flamingo, KOSMOS, llava} that employs various techniques to align visual signals obtained from the vision encoder to the language semantic space and input vision tokens and language tokens together into the large language decoder.

\begin{figure*}[t]
\centering
\includegraphics[width=1.0\linewidth]{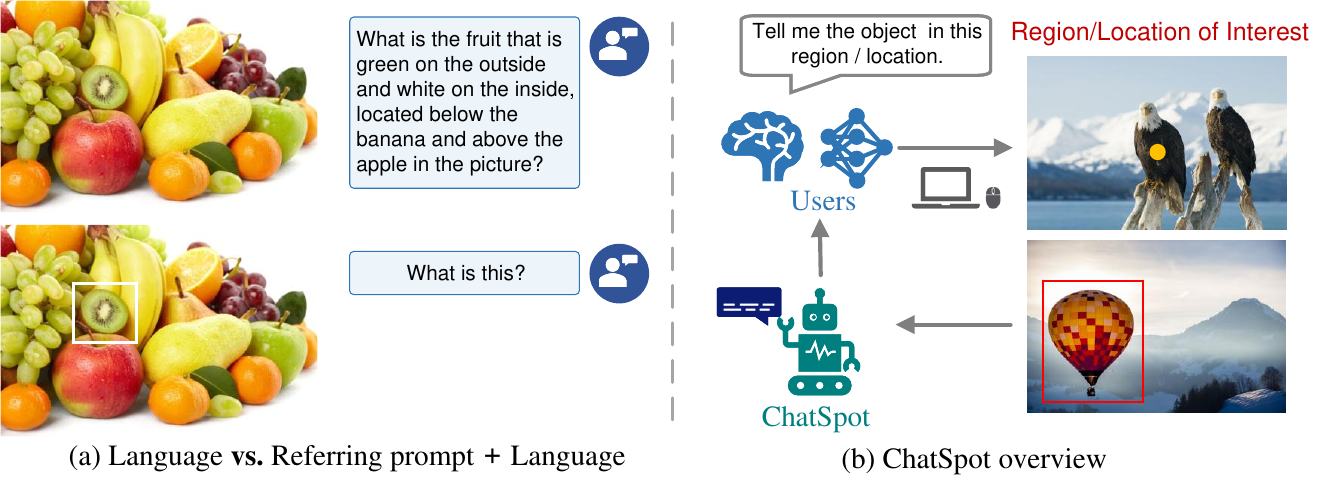}
\caption{(a) is the intuitive comparison between language instruction and the combination of region prompt and language instruction. (b) is the overview of ChatSpot. We extend the power of advanced LLM to vision-language modality and support a range of interaction forms including native language, mouse-clicking, and mouse boxing, enabling the interaction to be more flexible and user-friendly.}
\label{fig:1}
\vspace{-1.5mm}
\end{figure*}

Despite existing \textit{end-to-end} MLLMs have achieved remarkable progress in vision-language human-AI interaction, the mode of interactive instruction is still limited in language. When meeting complex scenes as illustrated in Figure \ref{fig:1} (a), it is difficult to only use the language to accurately describe the requirement of the user. However, if we can add some \textbf{referring prompts}, \eg, reference points, bounding boxes, \etc, to MLLMs, the model can focus on the region of interest (RoI) and achieve finer-grained interaction, which is more flexible and user-friendly. 

Motivated by this, we present ChatSpot, a fully end-to-end unified multimodal language model designed to empower special region vision-language interaction. As illustrated in Figure~\ref{fig:1} (b), ChatSpot extends the LLMs' power to incorporate diverse multimodal inputs, and it can support a range of interaction forms. Users can communicate with the system using their native language, as well as gestures such as clicking and drawing boxes that we call \textbf{\textit{Precise Referring}}, to obtain the desired information about the entire image or the region of interest (RoI). When a specific region is selected, ChatSpot can follow the precise referring instructions to perform various fine-grained applications, such as identifying jersey numbers or analyzing facial expressions in the given task, which is illustrated in Figure \ref{fig.4}. Furthermore, the precise referring can be regarded as a link of the chain-of-thought (CoT) to enhance the special logical reasoning ability of MLLMs. When an intelligent agent (robot or expert model) locates a target or region of interest based on user demands, ChatSpot can further analyze the details of this region and provide more specific suggestions and refined instructions, enabling the agent to interact with the physical world more effectively.

The success of ChatSpot hinges on three components: (1) We design a simple but effective precise referring instruction tuning method for MLLMs to support fine-grained interaction. (2) We construct a high-quality Multi-grained vision-language instruction-following dataset (\textbf{MGVLID}) including image-text and region-text with around $1.2M$ images and $3M$ query-answer pairs by collecting from existing datasets and generating based on GPT-4. (3) We design a series of evaluation tasks and metrics to assess the effectiveness of the proposed model. Extensive experiments have been conducted on a wide of vision-centric and vision-language benchmarks, and our ChatSpot shows excellent performance.

\section{Related Works}

\subsection{Large Language Models}
In recent years, large language models (LLMs) have garnered considerable attention in the domains of natural language processing (NLP) and artificial general intelligence (AGI) owing to their remarkable performance in language generation, in-context learning, world knowledge, and logical reasoning. Early works, \eg, BERT~\cite{Bert}, GPT-2~\cite{GPT-2} and T5~\cite{T5} established the foundation architecture of LLMs. Then, with the release of GPT-3~\cite{GPT3}, the first-ever language model to reach the parameter size of $175$ billion, LLM achieved impressive zero-shot performance on various language benchmarks. Furthermore, researchers discovered \textit{emergent ability}~\cite{wei2022emergent} in LLMs. That is when the model size of language models scales up to a certain level, there is a qualitative leap in the capabilities of language models. Sequentially, by aligning pre-trained GPT-3 to follow human language instructions through Reinforcement Learning with Human Feedback (RLHF)~\cite{christiano2017deep}, InstructGPT~\cite{InstructGPT} and ChatGPT~\cite{ChatGPT} showcased powerful capabilities for human-AI interaction, which make LLMs reach its ``iPhone moment''. Inspired by the great success of GPT series, many other open-sourced LLMs, such as OPT~\cite{OPT}, LLaMA~\cite{llama}, and GLM~\cite{GLM}, have been proposed, which achieve similar performance to GPT-3. Based on these open-sourced LLMs, several specific fine-tuned models are proposed to construct LLMs for various applications. For instance, Alpaca~\cite{alpaca} proposes a self-instruct framework based on LLaMA~\cite{llama} and employs $52K$ instructions generated by ChatGPT~\cite{ChatGPT} to construct an exceptional dialogue model.

\subsection{LLM-based MultiModal Interactive Agent}
The success of LLMs~\cite{ChatGPT, llama} may have opened the gate towards artificial general intelligence (AGI), a crucial component of which is human-AI interaction. The powerful zero-shot and logical reasoning ability of LLM makes it the central controller of the interactive system to schedule various application tools for different modality tasks, such as VQA, image editing, and image captioning. There are two mainstream interactive styles, \ie, plugin-based and end-to-end interaction. Plugin-based methods~\cite{VisualChatGPT, MMREACT, Hugginggpt, taskmatrix, yang2023gpt4tools} usually prompt LLM (ChatGPT~\cite{ChatGPT}, GPT-4~\cite{GPT4} or LLaMA~\cite{llama}) to invoke different plugins from other foundation or expert models to perform specific functions according to human instructions. However, despite plugin-based methods that enable diverse applications, they are limited in the effectiveness of plugin invocation and the performance of the plugin model. On the contrary, end-to-end interactive systems usually use a single large multimodal model to accomplish interaction. This approach takes advantage of cross-modal transfer, aligning multimodal domains to a common language semantic space and then using autoregressive language models as decoders to output the language. Following this pipeline, Flamingo~\cite{Flamingo} developed a gated cross-attention trained on billions of image-text pairs to align vision and language modality, which shows strong performance in few-shot learning. BLIP-2~\cite{BLIP2} introduced Q-Former to align visual features with language space more effectively. More recently, LLaVA~\cite{llava} proposed to use a simple linear layer to replace Q-Former and design a two-stage instruction-tuning procedure. Although existing end-to-end methods achieve remarkable performance in high efficiency, they are all limited to the interaction form of the full image and language-only instruction, which can not satisfy the demand for the specific region interaction. In this work, we build an end-to-end unified multimodal language model that supports a range of interaction forms that supports both full images and specific region.

\section{Methods}

ChatSpot is a multimodal large language model capable of perceiving real-world multimodal information, as well as following instructions, reasoning, and interacting with humans in natural language. It supports diverse forms of interaction including natural language, mouse-clicking, and mouse boxing. In this work, we mainly consider the modalities of image and language. And we will support more diverse interaction modalities and forms, \eg, video and audio, in the future.

\subsection{Overall Architecture}

\begin{figure*}[t]
\centering
\includegraphics[width=1.0\linewidth]{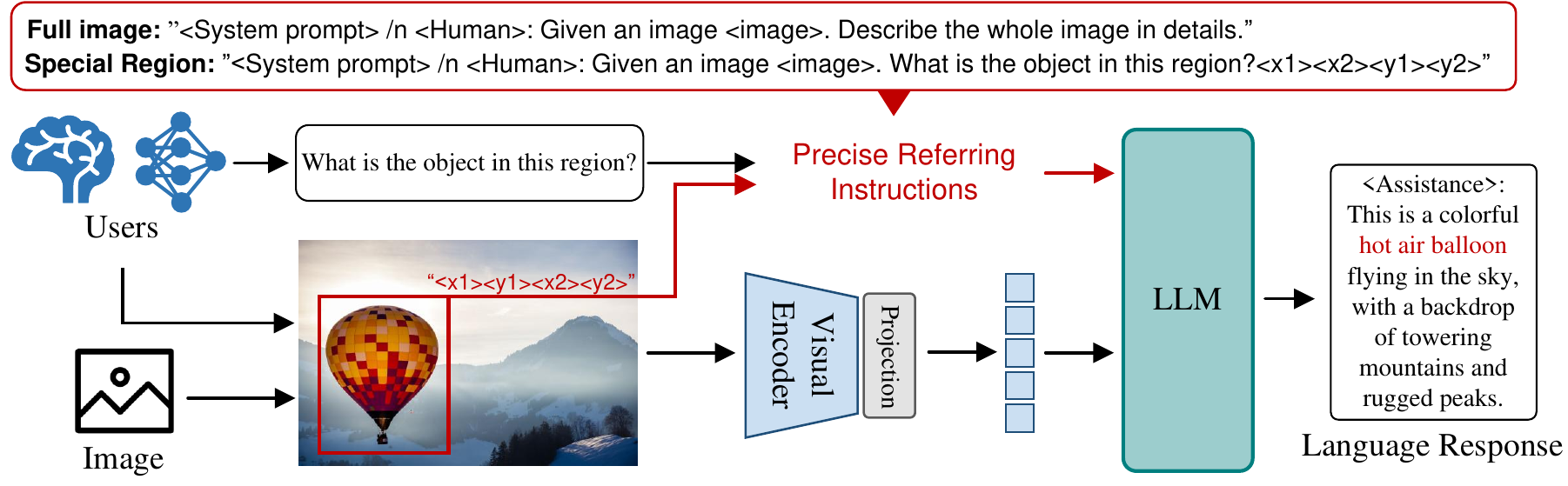}
\caption{\textbf{Overall pipeline of ChatSpot}. The architecture of ChatSpot consists of three main components: (1) an image encoder, (2) a large language model, and (3) a modality-align projector.}
\label{fig.2}
\end{figure*}

As illustrated in Figure \ref{fig.2}, ChatSpot consists of an image encoder, and a decoder-only LLM, and a modality alignment block. Inspired by LLaVA~\cite{llava}, ChatSpot incorporates a simple multilayer perceptron (MLP) to align the visual tokens with the space of language. The overall architecture is simple and does not use any extra AI models or post-processing operations. Different from previous end-to-end interactive systems~\cite{llava, minigpt4} that only support full image interaction, ChatSpot presents a more flexible interaction that supports users in further selecting the region of interest (RoI) to issue finer-grained instructions. 

When an image $I$ is uploaded by users to the ChatSpot system, users can use a mouse to select the RoI (points or boxes) $R_t$ through a series of gestures, such as clicking and drawing boxes, and give some language instructions $X^{\texttt{instruct}}$ about these RoIs. The system then converts the position of $R_t$ into the region prompt and connects it with the $X^{\texttt{instruct}}$ to generate the precise referring instructions. Afterward, the image $I$ is inputted into the visual encoder to extract visual tokens. And then the modality-align projector transforms the visual tokens to the language semantic space. After obtaining the refined visual tokens and precise referring instructions. LLM decoder $\mathcal{F}$ takes them as inputs and generates the response language sequence $Y$ autoregressively. Formally,
\begin{align}
\begin{aligned}
& V = \zeta \circ \mathcal{G}(I), \\
& Y = \mathcal{F}\Big(V, {\Phi(R_{t}), X^{\texttt{instruct}}}\Big),
\end{aligned}
\end{align}
where $V$ are aligned visual tokens. $\mathcal{G}$ is the visual encoder and $\zeta$ denotes the vision-language alignment projection. $\mathcal{F}$ is the large language decoder. $\Phi(\cdot)$ is the normalization operation. 

\subsection{Precise Referring Instruction}

Due to the inherent semantic unit mismatch between images and texts, it is ineffective to directly use the whole image and language sentences to describe the vision-language task. To this end, we propose \textit{precise referring instruction} that enables the unification of multi-grained vision-language task descriptions and supports proxy interaction forms. Specifically, we divide the instructions into two types, \ie, image-level instructions and region-level instructions.

\textbf{Image-level Instructions.} The image-level instructions are usually used to describe the task of the whole image, and existing multimodal instructions mostly adopt this form. For instance, given an image and we want to know what the content of the image is. Then the instruction can be like \textit{``Given an image \texttt{<image>}. Describe the whole image in detail''}, or \textit{``Given an image \texttt{<image>}, please tell me: \texttt{<question>}''}, where \textit{\texttt{<image>}} is the input image and \textit{\texttt{<question>}} denote some relative questions about images. Afterward, the LLM ingests the whole sentence and outputs the corresponding response.

\textbf{Region-level Instructions.} Compared to the overall information about the whole image, we often pay more attention to the information in specific regions. Therefore, it is valuable to design effective region-level instructions for multimodal large language models. The key challenge of region-level instructions is how to make LLM aware of the specific location of the region of interest. Here, we provide a simple but effective instruction format to achieve it. Specifically, we first define a unified region representation format as a tuple $R_t = \{x_k, y_k\}_{k=1}^{N}$ that represents $N$ points located in the selected region. Then the coordinates of selected points are normalized to $[0, 1]$ and transferred to the text tokens as $\Phi(R_{t})$. Finally, the region coordinate tokens are connected with the language instructions to generate the final region-level instructions. (~\cite{llava} and~\cite{sparksAGI} have served as evidence that LLM possesses the capability to comprehend spatial relationships and coordinates based on textual descriptions.) A simple example of region-level instruction is as follows: \textit{``Given an image \texttt{<image>}, What is the object doing in the region? \texttt{<region>}''} where \textit{\texttt{<region>}} = $\texttt{<box>}\Phi(R_{t})\texttt{</box>}$, $\texttt{<box>}$ and $\texttt{</box>}$ are special tokens to tell the LLM that this is a set of coordinates of the RoI.

Notably, the number of selected points $N$ is set freely so that we can achieve multi-grained interaction, such as points, boxes, and polygons.

\begin{figure*}[t]
\centering
\includegraphics[width=1.0\linewidth]{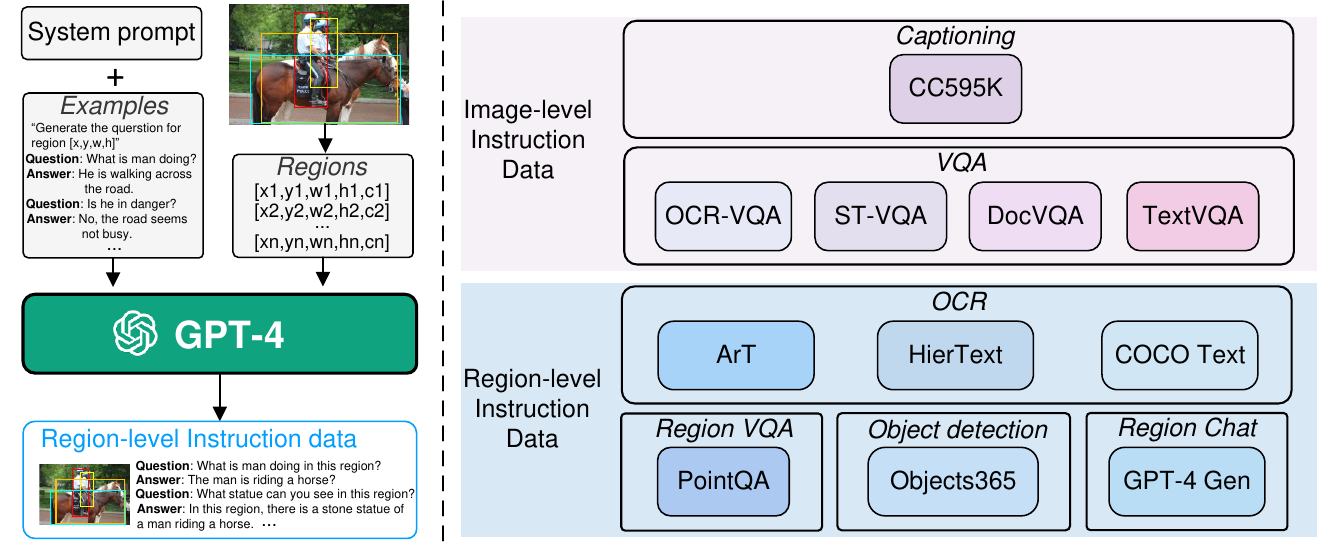}
\caption{Illustration of the pipeline to collect region-level chatting data for MGVILD (left) and dataset groups included in Multigrained Vision-language Instruction Datasets, \textbf{MGVLID} (right).}
\label{fig.3}
\end{figure*}

\subsection{Multi-grained Vision-language Instruction-following Dataset}

In order to empower ChatSpot with the precise referring instruction following ability, we need to construct enough high-quality region-level instruction datasets. However, existing vision-language datasets lack diverse region-level chatting data
. To this end, we design a data collection pipeline with the assistance of GPT-4~\cite{GPT4}, as shown in Figure~\ref{fig.3}. Inspired by LLaVA~\cite{llava}, we use the captions and bounding boxes of the target as the prompts and leverage GPT-4 to refine these captions and generate more relative and diverse conversation data. The difference is that our approach distinguishes itself by enforcing the alignment of every generated dialogue with precise regional coordinates. To achieve this, we leverage the VisualGenome dataset~\cite{krishna2017visual}, which provides comprehensive annotations of objects, attributes, and relationships within each image, enabling us to construct region-level instruction following datasets. These datasets consist of dense region-wise captions organized alongside carefully curated seed examples, which are used to query GPT-4 in an in-context-learning fashion. Through this pipeline, we have successfully gathered a total of $108K$ region-level instruction following samples.

Based on this data generation pipeline, we build a high-quality \textbf{M}ulti-\textbf{G}rained \textbf{V}ision-\textbf{L}anguage \textbf{I}nstruction-following \textbf{D}ataset, named \textbf{MGVLID}. MGVLID consists of two main parts, \ie, image-text instruction-following data and region-text instruction-following data. The former data consists of the image and the caption of the entire image, while the latter consists of the image, the bounding boxes of the target in the image, and the corresponding target captions. As shown in Figure~\ref{fig.3}, the whole MGVLID covers $11$ source datasets and we hold out $4$ datasets for model evaluation purposes.

\textbf{Image-level Instruction Data.} To gather image-level instruction-following data, we collect a wide range of publicly available multimodal datasets that have been human-annotated. We then transform these datasets into a unified instruction-following format. Specifically, we assemble a plethora of commonly used Question-Answering (QA), captioning, and object detection datasets, including CC595K (filtered based on CC3M~\cite{CC3M}), OCR-VQA~\cite{OCRVQA}, ST-VQA~\cite{STVQA}, DocVQA~\cite{DocVQA}, TextVQA~\cite{TextVQA} and Object365~\cite{objects365}. For each dataset, we design a series of unique instruction templates. These templates are subsequently carefully filtered and refined manually to ensure optimal rationales and diversity of the conversation. Due to the considerable differences in label lengths among various task datasets (such as caption or category words), we incorporate additional instruction tags to specify the desired response style. For instance, we include tags like \textit{``answer in shot''} for short-answer data and \textit{``answer in detail''} for long-answer data.

\textbf{Region-level Instruction Data.} While image-level instruction data empowers the model's global visual perception and human instruction-following ability, region-level instruction data offers region-level observation and more fine-grained instructions, enabling the model to further acquire spatial perception and reasoning abilities. In order to construct region-level instruction datasets, we first collect region-text pairs based on existing region-level task (object detection and OCR) datasets, \ie, Object365~\cite{objects365}, COCO text~\cite{coco_text}, HierText~\cite{hiertext} and Art~\cite{ArT}. We collected region-text pairs that consist of instance-level bounding boxes and their corresponding content. Subsequently, we utilize unique instruction templates to further refine these region-text pairs, resulting in a series of questions and answers. Furthermore, we also collect the PointQA datasets from LookTwice-QA~\cite{PointQA} to support point-wise referring instruction tuning, where the models are asked to answer questions based on the input points or boxes. By incorporating the high-quality dense region chatting data generated based on GPT-4, the final region-level instruction data is constructed.

\section{Experiments}
\label{exp}

\subsection{Implementation Details}

To build ChatSpot, we implement the CLIP ViT-L/14~\cite{radford2021learning} as the visual encoder to encode images. For the large language model, we choose open-sourced Vicuna-7B~\cite{vicuna} as the language decoder, a LLaMA model fine-tuned with instructions. For alignment projection, we just adopt a simple linear layer to connect vision and language embedding space.

Inspired by LLaVA, the model is trained in a two-stage fashion. Firstly, we initialize the model using pre-trained weights from LLaMA and CLIP ViT. During this first stage, we only train the projection layer. Meanwhile, we freeze the majority of the LLM parameters. In this stage, we mainly use the image-text instruction-following data of MGVLID to train the model for vision-language instruction-following alignment learning. In the second stage, we only freeze the visual encoder and unfreeze the LLM parameters. In this stage, we mainly use the region-text instruction-following data including RegionChat to train the model for region-level instruction-following and multi-turn chatting ability. Specifically, the model is fine-tuned over $3$ epochs, with a batch size of $128$. AdamW~\cite{AdamW} optimizer is employed, and the learning rate is set to $2e-3$ in  the first training stage and $2e-5$ in the second training stage. For LLM, the maximum length of tokens is set to $2,048$.

\subsection{Task Evaluation}

In order to objectively showcase ChatSpot's region recognization and zero-shot ability. We choose several downstream tasks including regional classification, OCR text recognization, and VQA answer grounding tasks to evaluate ChatSpot. The results are shown in Table \ref{table:det} and Table \ref{table:qa}. Notably, all the experiments of different tasks are conducted by the shared-parameter generalist model, and we just change the language instructions for different tasks. 

\textbf{Regional Classification.} Object detection is a fundamental vision task that consists of object location and recognition subtasks. In this part, we mainly evaluate the regional classification ability of ChatSpot in COCO~\cite{COCO}, which is a common dataset in object detection tasks. Specifically, we first use the GT boxes or the bounding boxes generated by existing detectors, such as DINO~\cite{zhang2022dino}, as region prompts to ask ChatSpot to answer what category it is. For an example, we use \textit{``What can you see in this region? \texttt{<region>}''}, where \textit{\texttt{<region>}} denotes the coordinates of the region boxes. Then we compute the metrics of Average Precision (AP) and Accuracy about the bounding boxes with the predicted classes. Notably, due to ChatSpot's output typically being a single sentence, it cannot be directly used as a category for evaluation. Here, we employ the CLIP text encoder to calculate the text feature similarity between the output sentence and all COCO categories for category determination.

As shown in Table \ref{table:det}, we randomly select $1,000$ images from the COCO validation set, namely COCO-$1000$, to evaluate ChatSpot for efficiency. Our ChatSpot achieves $64.5\%$ accuracy on COCO-$1000$ with the provided GT boxes. When given DINO-generated bounding boxes as region prompts, our ChatSpot achieves $39.6\%$ AP, which is also a competitive performance. Notably, We do not use any annotations from COCO. Therefore, the results show that ChatSpot achieves impressive zero-shot classification ability in region-level recognization. 

\textbf{Regional Optical Character Recognition.} Optical Character Recognition (OCR) is a visual entity recognition task that requires the recognition of the graphemes in a written text. In this part, we select COCO text~\cite{coco_text} to evaluate the regional text recognization ability of ChatSpot. COCO Text is a large-scale dataset for text detection and recognition. We first use the provided region boxes of the datasets as the region referring and ask the ChatSpot \textit{``What text can you see in this region?''\texttt{<region>}}. Then ChatSpot will respond to the specific answer. Similar to evaluating VizWiz, when the answered sentence includes the correct GT answer, we consider ChatSpot's response to be correct. As shown in Table \ref{table:qa}, our ChatSpot achieves $31.8\%$ accuracy on the COCO Text validation set. 

\textbf{Visual Question Answering (VQA).} VQA is the task of answering open-ended questions based on the whole image or the region of interest (RoI), which is well-suited for evaluating the perceptual and reasoning abilities of large multimodal language models in understanding image content. In this part, we choose two datasets to evaluate ChatSpot. One is VizWiz-VQA-Grounding dataset~\cite{VizWiz}, a dataset that visually grounds answers to visual questions asked by people with visual impairments. Another is PointQA~\cite{PointQA}, a set of datasets that require a pointer to an object in the image to be answered correctly. These two datasets both provide specific regions (boxes or points) and ask the question about the corresponding area. Therefore, it also requires the model to possess the ability of region-level perception and reasoning ability. Specifically, given a region of interest and the corresponding question, we first input the coordinate of region boxes to ChatSpot with the corresponding question. If the output answer sentence of ChatSpot includes the GT answer, we consider the response to be correct. As shown in Table~\ref{table:qa}, benefiting from ChatSpot's strong region-level instruction following abilities, the model achieves competitive performance that obtains $63.0\%$ accuracy on VizWiz and $68.2\%$ accuracy on PointQA (boxes given) and $62.0\%$ on PointQA (points given).

\begin{figure*}[hbtp]
\centering
\includegraphics[width=1.0\linewidth]{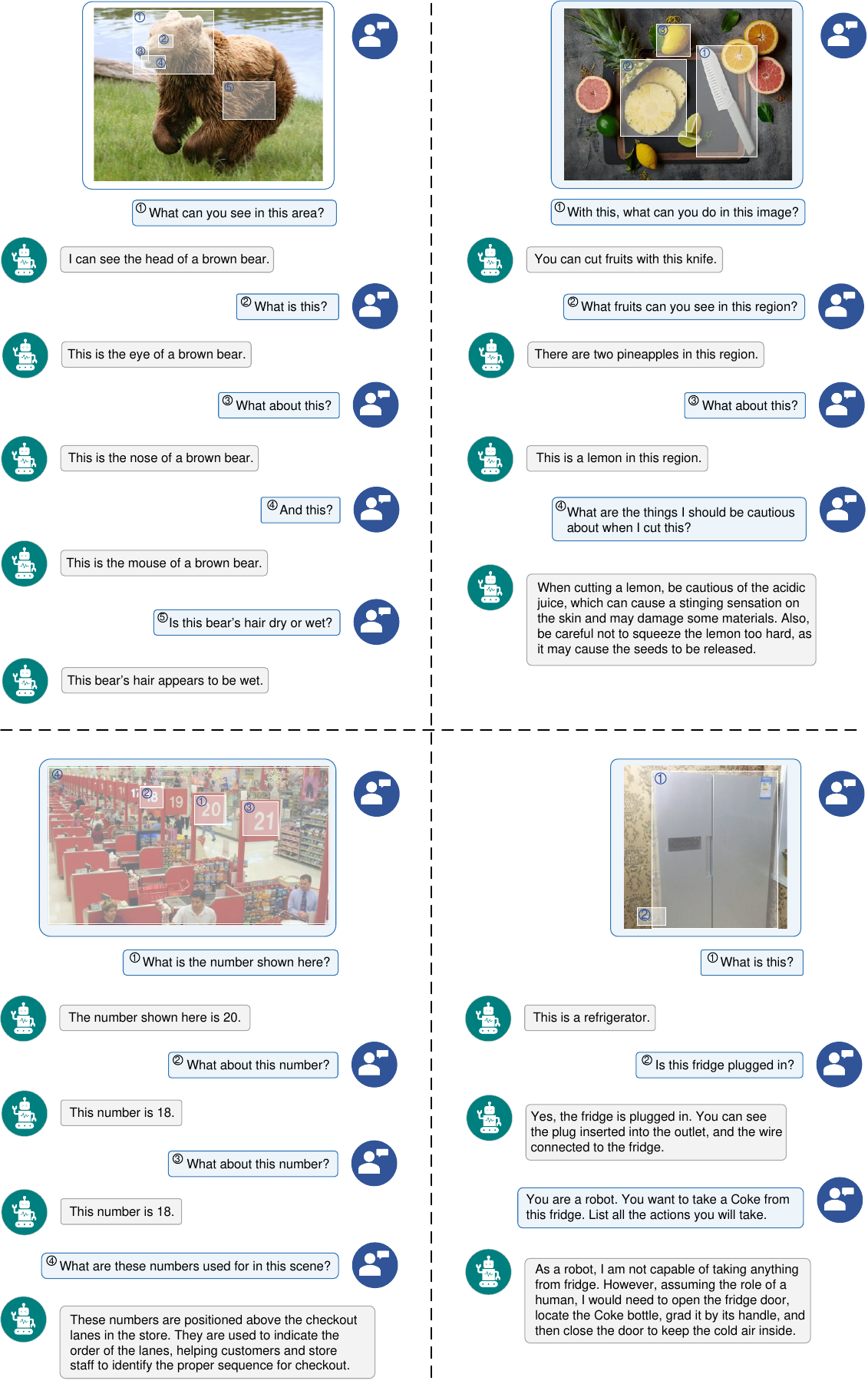}
\caption{Cases of interacting with ChatSpot through drawing bounding boxes.}
\label{fig.4}
\end{figure*}

\begin{table}[t] 
\begin{center}
\caption{\textbf{Zero-shot region recognition results on COCO \texttt{val} set.} We randomly select $1,000$ images from the COCO validation set for evaluation. The referring regions are provided by GT boxes and advanced detector DINO~\cite{zhang2022dino}. Acc. denotes the classification accuracy when given the GT boxes.}

\setlength{\tabcolsep}{0.57mm}
\resizebox{0.9\columnwidth}{!}{
\begin{tabular}{lccccccccccccc}
      \toprule
      \multirow{2}{*}{Method} & \multirow{2}{*}{Backbone} & 
       \multirow{2}{*}{\begin{tabular}[c]{@{}c@{}}Training Data\end{tabular}} & \multirow{2}{*}{Region} &
      \multicolumn{8}{c}{COCO \cite{COCO}} \\
      \cmidrule(lr){5-12}
      &  &  &  & \multicolumn{2}{c}{AP} & \multicolumn{1}{c}{AP$\rm_{50}$}  & \multicolumn{1}{c}{AP$\rm_{75}$} & \multicolumn{1}{c}{AP$\rm_{\texttt{s}}$} & \multicolumn{1}{c}{AP$\rm_{\texttt{m}}$}  & \multicolumn{1}{c}{AP$\rm_{\texttt{l}}$} & \multicolumn{1}{c}{Acc.}  \\
      \hline
      \multicolumn{10}{l}{\small{\textbf{\emph{Multi-modal Large Language Models}}}}  \\
      ChatSpot & CLIP-ViT & MGVLID  & GT boxes & \multicolumn{2}{c}{48.8} & 48.8 & 48.8 & 35.1 & 56.0 & 60.3 & 64.5 \\
      ChatSpot & CLIP-ViT & MGVLID  & DINO boxes & \multicolumn{2}{c}{39.6} & 50.2 & 44.1 & 21.6 & 45.8 & 58.8 & - \\
\bottomrule
\end{tabular}

}
\label{table:det}
\end{center}
\end{table}

\begin{table}[t] 
\begin{center}
\caption{\textbf{Experimental results on a diverse set of downstream tasks.} We also evaluate ChatSpot on a series of downstream tasks including optical character recognition (OCR) and visual question answering (VQA). We mainly report the metric of Accuracy ($\%$) for evaluation. For PointQA, ``B'' and ``P'' mean that answer the question based on the given box and point, respectively.}

\setlength{\tabcolsep}{0.57mm}
\resizebox{0.88\columnwidth}{!}{
\begin{tabular}{lccccccccccc}
      \toprule
      \multirow{2}{*}{Method} & \multirow{2}{*}{Backbone} &
       \multirow{2}{*}{\begin{tabular}[c]{@{}c@{}}Training Data\end{tabular}} &
      \multicolumn{2}{c}{OCR} & \multicolumn{6}{c}{VQA} \\
      \cmidrule(lr){4-5} \cmidrule(lr){6-11} 
      &  & &\multicolumn{2}{c}{COCO Text} & \multicolumn{2}{c}{VizWiz}  & \multicolumn{2}{c}{PointQA (B)} & \multicolumn{2}{c}{PointQA (P)} \\
      \hline
      ChatSpot & CLIP-ViT& MGVLID & \multicolumn{2}{c}{31.8} & \multicolumn{2}{c}{63.0} & \multicolumn{2}{c}{68.2}  & \multicolumn{2}{c}{62.0}\\
\bottomrule
\end{tabular}
}
\label{table:qa}
\end{center}
\end{table}

\subsection{Qualitative Analysis}

In order to provide a comprehensive showcase of ChatSpot, we selected several classic cases to demonstrate its specific abilities. We specifically demonstrate four core capabilities of ChatSpot through these examples as follows:

\textbf{Region Perception Ability.} As shown in Figure \ref{fig.4} (left top), we first show the case that depicts ChatSpot's ability to perceive the region of interest and recognize the corresponding context. In this case, ChatSpot can identify the selected area of different levels of granularity, \eg, the head of the brown bear and the nose of the brown bear. It can also perceive some specific features of the target like the wet hair of the bear. By collaborating with human referring prompts (point or boxes), ChatSpot showcases its powerful capability in perceiving details, which provides sufficient detailed information for robots to perform more refined operations.

\textbf{Content Generation Capability.} ChatSpot also possesses a powerful content generation capability related to regions of interest, as illustrated in Figure \ref{fig.4} (left-right). In this case, ChatSpot first recognizes the corresponding object (lemon) in the region of interest. Then it can also generate responses for other information that cannot be captured in the image, \ie, the precautions when cutting a lemon with a knife. Thanks to the huge knowledge of LLMs, ChatSpot can provide rich content explanations or suggestions based on visual information.

\textbf{Optical Character Recognition (OCR) Ability.} ChatSpot also achieves impressive performance in recognizing the optical character, \eg, text, number, and signal. As shown in Figure \ref{fig.4} (bottom left), ChatSpot can accurately identify the number written on the signboard and analyze the purpose of these numbers based on the global context information.

\textbf{Special Reasoning Ability.} In addition to perception, another important capability of ChatSpot is spatial reasoning which can further analyze the region of interest based on its knowledge after recognition, which is important for robotics and automation. As shown in Figure \ref{fig.4} (bottom right), ChatSpot first identifies the fridge and then determines that the refrigerator is powered on according to the power wine being plugged into the power strip and connected to the refrigerator. Furthermore, ChatSpot provides a series of detailed action instructions when the user wants to take a Coke from the fridge. This enables the robots to be able to make further decisions regarding fine-grained operations after inferring the specific status (fridge is plugged in.) of the RoI.

\section{Discussion}
\label{discussion}

In this section, we are dedicated to delving deeper into the capabilities and features of ChatSpot, as well as identifying the limitations that currently hinder further enhancement of ChatSpot's abilities.

\subsection{Robustness on Region Referring}
\begin{figure}
\centering
    \begin{minipage}{0.6\textwidth}
    \centering
	\subfigure[]{
        \includegraphics[width=0.45\columnwidth]{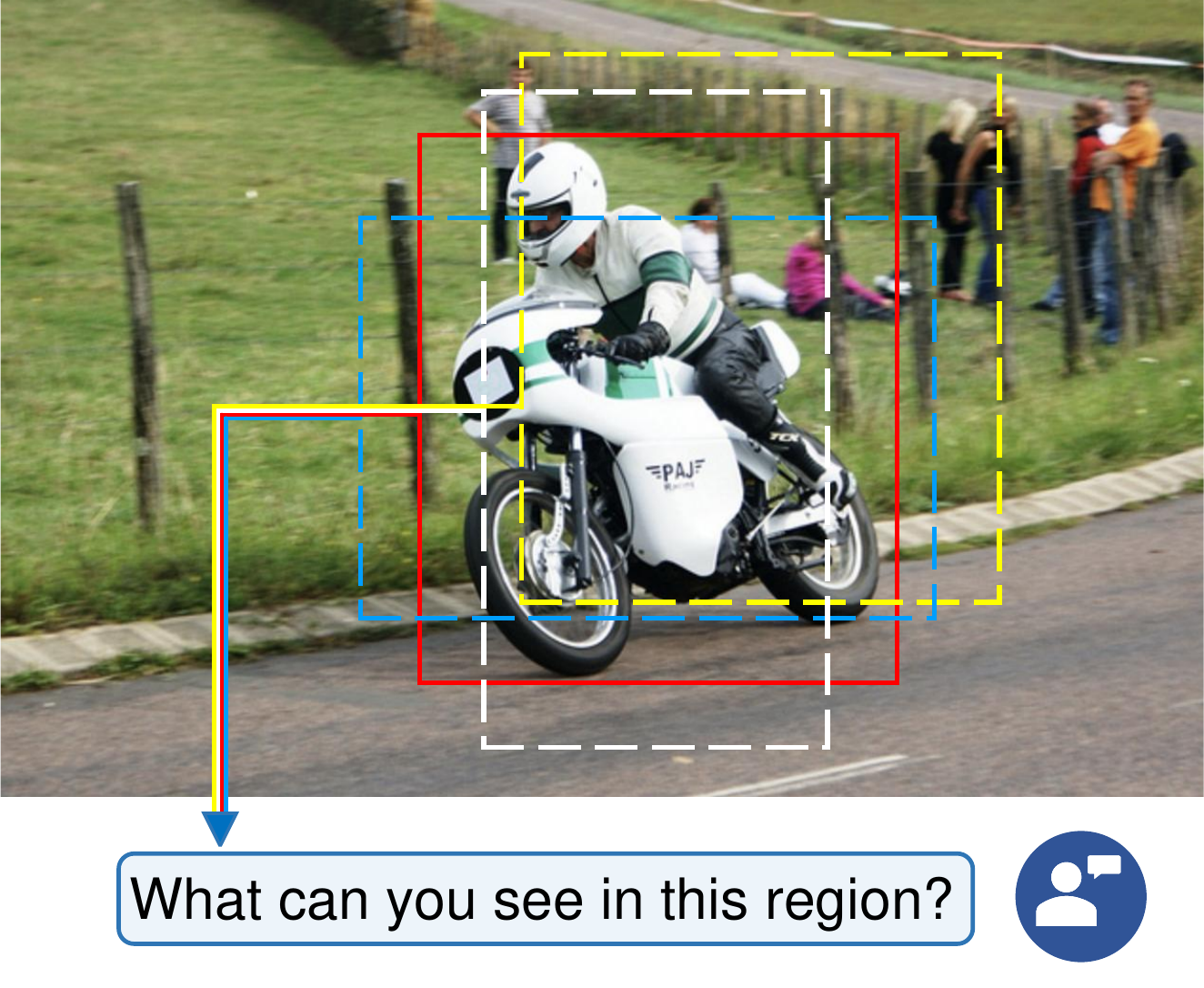}
    }
    \subfigure[]{
	\includegraphics[width=0.47\columnwidth]{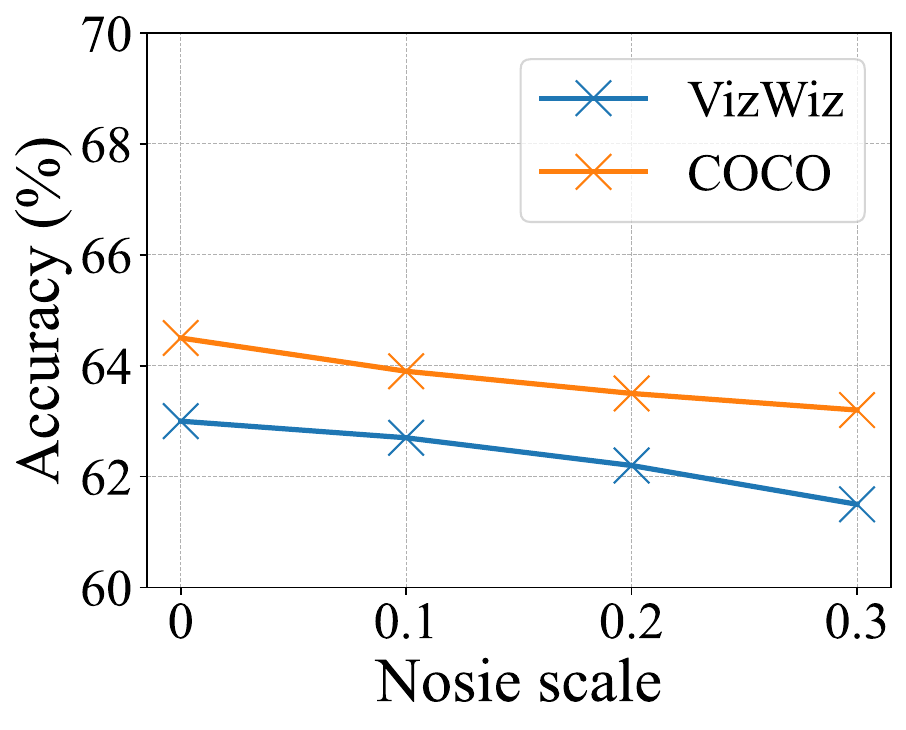}
    }
	\end{minipage}
	\begin{minipage}{0.3\textwidth}
	\caption{\label{fig:5}
	\textbf{Experiment on the robustness of region referring.} (a) demonstrates the process of randomly adding noise to the original region boxes and posing the question to ChatSpot. (b) showcases the performance of ChatSpot after incorporating random noise perturbations.}
    \end{minipage}
\end{figure}

The process of selecting regions, whether through drawing boxes or clicking, plays a crucial role in ChatSpot. However, users often face difficulties in accurately annotating their areas of interest. In such instances, it is essential for ChatSpot to exhibit a high level of robustness in region selection. Hence, an analysis of the robustness of ChatSpot in the region referring is conducted.
Specifically, we randomly introduce box noises of different scales (scale = $0.1$, $0.2$, and $0.3$) to the region boxes of the COCO and VizWiz as illustrated in Figure \ref{fig:5} (a). Figure \ref{fig:5} (b) demonstrates that the performance of ChatSpot on COCO and VizWiz does not show a significant decrease after introducing box noises into the region bounding box, which means ChatSpot possesses strong robustness in region referring.

\subsection{Region Referring Hallucination}

There exists a potential risk that ChatSpot may mistakenly recognize the region referred to by users as a nearby region, and we coin this phenomenon as the ``region referring hallucination''. To quantitatively assess whether ChatSpot is prone to region referring hallucination issues, we analyze the cases of misidentification in the sampled COCO-$1000$ dataset. Specifically, we define an occurrence of the ``region referring hallucination'' when an object of a certain class, misidentified by ChatSpot, appears in close proximity to the selected region, with an Intersection over Union (IoU) greater than $0.5$.  

\begin{table}[ht] 
\begin{center}
\caption{\textbf{Results of region referring hallucination.} $s$ is the noise scale. ``Acc.'' indicates the Accuracy and ``Hallucination Ratio'' denotes the proportion of region referring hallucination.}

\setlength{\tabcolsep}{0.5mm}
\resizebox{0.5\columnwidth}{!}{
\begin{tabular}{llccccc}
\toprule
\multirow{2}{*}{\textbf{Model}} & \multirow{2}{*}{\textbf{Setting}} & \multicolumn{2}{c}{\textbf{COCO}} \\
 &  & Acc. & Hallucination Ratio \\ \midrule
\multirow{4.5}{*}{ChatSpot{}} & No noise & 64.7 & 2.2 \\
& Box noise ($s=0.1$) & 63.9 & 2.4 \\
& Box noise ($s=0.2$) & 63.5 & 2.1 \\
& Box noise ($s=0.3$) & 63.2 & 2.3 \\
\bottomrule
\end{tabular}
}
\label{table:ill}
\end{center}
\end{table}

As shown in Table \ref{table:ill}, ChatSpot has shown very few instances of regional illusions, accounting for only about $2\%$ of the total. Furthermore, even with the introduction of box noise, the illusion ratio does not have a significant change ($2.2\%$-$2.3\%$). This result demonstrates that ChatSpot exhibits a capability for precise region referring. The current recognition errors can be largely attributed to the model's insufficient training on an adequate amount of data.

\subsection{Limitations}
Although ChatSpot has achieved remarkable performance in precise region referring and special region reasoning, it still has some noticeable limitations. For example, ChatSpot currently lacks the ability to support referential output boxes and does not support the recognition of certain special symbols, such as license plate numbers. These shortcomings can be attributed to insufficient training data. Another limitation is the phenomenon of catastrophic forgetting, whereby fine-tuning ChatSpot on a new dataset leads to the forgetting of previously acquired knowledge, leading to the overall performance bottleneck. Furthermore, it is hard to evaluate the region recognition ability of multi-modal large language models since its evaluation is
essentially different from traditional visual-language models. Though we take the first step to
quantitatively evaluate it by giving pre-defined boxes, it is still an open problem: \textit{how can we establish a comprehensive and automatic benchmark to evaluate existing multimodal large language models?} These limitations require further research from the community. ChatSpot currently only supports the interaction forms of mouse-clicking and drawing boxes. In the future, we will support more diverse interaction forms, \eg, polygon, and mask.

\section{Conclusion}

In this work, we first propose \textit{precise referring instruction} tuning for multimodal LLMs (MLLMs) that utilizes diverse reference representations for referring special regions. Based on precise referring instruction, we build ChatSpot, a fully end-to-end MLLM that supports diverse region referring prompts, \ie, points, and boxes. Then we construct a large-scale multi-grained vision-language instruction-following dataset, MGVLID. Trained on MGVLID, ChatSpot demonstrates outstanding performance both in interactive chatting and downstream tasks. Results suggest that combining precise referring instructions with MLLMs stimulates the model's ability for special region understanding and reasoning. We hope this work can spur more advanced MLLMs in the future.

{
\small
\bibliographystyle{splncs04}
\bibliography{egbib}

\begin{thebibliography}{10}
\providecommand{\url}[1]{\texttt{#1}}
\providecommand{\urlprefix}{URL }
\providecommand{\doi}[1]{https://doi.org/#1}

\bibitem{Flamingo}
Alayrac, J., Donahue, J., Luc, P., Miech, A., Barr, I., Hasson, Y., Lenc, K.,
  Mensch, A., Millican, K., Reynolds, M., Ring, R., Rutherford, E., Cabi, S.,
  Han, T., Gong, Z., Samangooei, S., Monteiro, M., Menick, J.L., Borgeaud, S.,
  Brock, A., Nematzadeh, A., Sharifzadeh, S., Binkowski, M., Barreira, R.,
  Vinyals, O., Zisserman, A., Simonyan, K.: Flamingo: a visual language model
  for few-shot learning. In: NeurIPS (2022)

\bibitem{STVQA}
Biten, A.F., Litman, R., Xie, Y., Appalaraju, S., Manmatha, R.: Latr:
  Layout-aware transformer for scene-text vqa. In: Proceedings of the IEEE/CVF
  conference on computer vision and pattern recognition. pp. 16548--16558
  (2022)

\bibitem{GPT3}
Brown, T., Mann, B., Ryder, N., Subbiah, M., Kaplan, J.D., Dhariwal, P.,
  Neelakantan, A., Shyam, P., Sastry, G., Askell, A., et~al.: Language models
  are few-shot learners. Advances in neural information processing systems
  \textbf{33},  1877--1901 (2020)

\bibitem{sparksAGI}
Bubeck, S., Chandrasekaran, V., Eldan, R., Gehrke, J., Horvitz, E., Kamar, E.,
  Lee, P., Lee, Y.T., Li, Y., Lundberg, S., et~al.: Sparks of artificial
  general intelligence: Early experiments with gpt-4. arXiv preprint
  arXiv:2303.12712  (2023)

\bibitem{vicuna}
Chiang, W.L., Li, Z., Lin, Z., Sheng, Y., Wu, Z., Zhang, H., Zheng, L., Zhuang,
  S., Zhuang, Y., Gonzalez, J.E., Stoica, I., Xing, E.P.: Vicuna: An
  open-source chatbot impressing gpt-4 with 90\%* chatgpt quality.
  \url{https://lmsys.org/blog/2023-03-30-vicuna/} (2023)

\bibitem{ArT}
Chng, C.K., Liu, Y., Sun, Y., Ng, C.C., Luo, C., Ni, Z., Fang, C., Zhang, S.,
  Han, J., Ding, E., et~al.: Icdar2019 robust reading challenge on
  arbitrary-shaped text-rrc-art. In: 2019 International Conference on Document
  Analysis and Recognition (ICDAR). pp. 1571--1576. IEEE (2019)

\bibitem{christiano2017deep}
Christiano, P.F., Leike, J., Brown, T., Martic, M., Legg, S., Amodei, D.: Deep
  reinforcement learning from human preferences. Advances in neural information
  processing systems  \textbf{30} (2017)

\bibitem{Bert}
Devlin, J., Chang, M.W., Lee, K., Toutanova, K.: Bert: Pre-training of deep
  bidirectional transformers for language understanding. arXiv preprint
  arXiv:1810.04805  (2018)

\bibitem{VizWiz}
Gurari, D., Li, Q., Stangl, A.J., Guo, A., Lin, C., Grauman, K., Luo, J.,
  Bigham, J.P.: Vizwiz grand challenge: Answering visual questions from blind
  people. In: Proceedings of the IEEE conference on computer vision and pattern
  recognition. pp. 3608--3617 (2018)

\bibitem{KOSMOS}
Huang, S., Dong, L., Wang, W., Hao, Y., Singhal, S., Ma, S., Lv, T., Cui, L.,
  Mohammed, O.K., Liu, Q., et~al.: Language is not all you need: Aligning
  perception with language models. arXiv preprint arXiv:2302.14045  (2023)

\bibitem{krishna2017visual}
Krishna, R., Zhu, Y., Groth, O., Johnson, J., Hata, K., Kravitz, J., Chen, S.,
  Kalantidis, Y., Li, L.J., Shamma, D.A., et~al.: Visual genome: Connecting
  language and vision using crowdsourced dense image annotations. International
  journal of computer vision  \textbf{123},  32--73 (2017)

\bibitem{BLIP2}
Li, J., Li, D., Savarese, S., Hoi, S.: Blip-2: Bootstrapping language-image
  pre-training with frozen image encoders and large language models. arXiv
  preprint arXiv:2301.12597  (2023)

\bibitem{taskmatrix}
Liang, Y., Wu, C., Song, T., Wu, W., Xia, Y., Liu, Y., Ou, Y., Lu, S., Ji, L.,
  Mao, S., et~al.: Taskmatrix. ai: Completing tasks by connecting foundation
  models with millions of apis. arXiv preprint arXiv:2303.16434  (2023)

\bibitem{COCO}
Lin, T., Maire, M., Belongie, S.J., Hays, J., Perona, P., Ramanan, D.,
  Doll{\'{a}}r, P., Zitnick, C.L.: Microsoft {COCO:} common objects in context.
  In: ECCV. pp. 740--755 (2014)

\bibitem{llava}
Liu, H., Li, C., Wu, Q., Lee, Y.J.: Visual instruction tuning (2023)

\bibitem{hiertext}
Long, S., Qin, S., Panteleev, D., Bissacco, A., Fujii, Y., Raptis, M.: Towards
  end-to-end unified scene text detection and layout analysis. In: Proceedings
  of the IEEE/CVF Conference on Computer Vision and Pattern Recognition. pp.
  1049--1059 (2022)

\bibitem{AdamW}
Loshchilov, I., Hutter, F.: Decoupled weight decay regularization. In: {ICLR}
  (2019)

\bibitem{PointQA}
Mani, A., Yoo, N., Hinthorn, W., Russakovsky, O.: Point and ask: Incorporating
  pointing into visual question answering. arXiv preprint arXiv:2011.13681
  (2020)

\bibitem{DocVQA}
Mathew, M., Karatzas, D., Jawahar, C.: Docvqa: A dataset for vqa on document
  images. In: Proceedings of the IEEE/CVF winter conference on applications of
  computer vision. pp. 2200--2209 (2021)

\bibitem{OCRVQA}
Mishra, A., Shekhar, S., Singh, A.K., Chakraborty, A.: Ocr-vqa: Visual question
  answering by reading text in images. In: 2019 international conference on
  document analysis and recognition (ICDAR). pp. 947--952. IEEE (2019)

\bibitem{ChatGPT}
OpenAI: Chatgpt. \url{https://openai.com/blog/chatgpt/} (2023)

\bibitem{GPT4}
OpenAI: Gpt-4 technical report (2023)

\bibitem{InstructGPT}
Ouyang, L., Wu, J., Jiang, X., Almeida, D., Wainwright, C.L., Mishkin, P.,
  Zhang, C., Agarwal, S., Slama, K., Ray, A., Schulman, J., Hilton, J., Kelton,
  F., Miller, L., Simens, M., Askell, A., Welinder, P., Christiano, P.F.,
  Leike, J., Lowe, R.: Training language models to follow instructions with
  human feedback. In: NeurIPS (2022)

\bibitem{radford2021learning}
Radford, A., Kim, J.W., Hallacy, C., Ramesh, A., Goh, G., Agarwal, S., Sastry,
  G., Askell, A., Mishkin, P., Clark, J., et~al.: Learning transferable visual
  models from natural language supervision. In: International conference on
  machine learning. pp. 8748--8763. PMLR (2021)

\bibitem{GPT-2}
Radford, A., Wu, J., Child, R., Luan, D., Amodei, D., Sutskever, I., et~al.:
  Language models are unsupervised multitask learners. OpenAI blog
  \textbf{1}(8), ~9 (2019)

\bibitem{T5}
Raffel, C., Shazeer, N., Roberts, A., Lee, K., Narang, S., Matena, M., Zhou,
  Y., Li, W., Liu, P.J.: Exploring the limits of transfer learning with a
  unified text-to-text transformer. The Journal of Machine Learning Research
  \textbf{21}(1),  5485--5551 (2020)

\bibitem{objects365}
Shao, S., Li, Z., Zhang, T., Peng, C., Yu, G., Zhang, X., Li, J., Sun, J.:
  Objects365: A large-scale, high-quality dataset for object detection. In:
  Proceedings of the IEEE/CVF international conference on computer vision. pp.
  8430--8439 (2019)

\bibitem{CC3M}
Sharma, P., Ding, N., Goodman, S., Soricut, R.: Conceptual captions: A cleaned,
  hypernymed, image alt-text dataset for automatic image captioning. In:
  Proceedings of the 56th Annual Meeting of the Association for Computational
  Linguistics (Volume 1: Long Papers). pp. 2556--2565 (2018)

\bibitem{Hugginggpt}
Shen, Y., Song, K., Tan, X., Li, D., Lu, W., Zhuang, Y.: Hugginggpt: Solving ai
  tasks with chatgpt and its friends in huggingface. arXiv preprint
  arXiv:2303.17580  (2023)

\bibitem{TextVQA}
Singh, A., Natarajan, V., Shah, M., Jiang, Y., Chen, X., Batra, D., Parikh, D.,
  Rohrbach, M.: Towards vqa models that can read. In: Proceedings of the
  IEEE/CVF conference on computer vision and pattern recognition. pp.
  8317--8326 (2019)

\bibitem{alpaca}
Taori, R., Gulrajani, I., Zhang, T., Dubois, Y., Li, X., Guestrin, C., Liang,
  P., Hashimoto, T.B.: Stanford alpaca: An instruction-following llama model.
  \url{https://github.com/tatsu-lab/stanford_alpaca} (2023)

\bibitem{llama}
Touvron, H., Lavril, T., Izacard, G., Martinet, X., Lachaux, M.A., Lacroix, T.,
  Rozi{\`e}re, B., Goyal, N., Hambro, E., Azhar, F., Rodriguez, A., Joulin, A.,
  Grave, E., Lample, G.: Llama: Open and efficient foundation language models.
  arXiv preprint arXiv:2302.13971  (2023)

\bibitem{coco_text}
Veit, A., Matera, T., Neumann, L., Matas, J., Belongie, S.: Coco-text: Dataset
  and benchmark for text detection and recognition in natural images. arXiv
  preprint arXiv:1601.07140  (2016)

\bibitem{wei2022emergent}
Wei, J., Tay, Y., Bommasani, R., Raffel, C., Zoph, B., Borgeaud, S., Yogatama,
  D., Bosma, M., Zhou, D., Metzler, D., et~al.: Emergent abilities of large
  language models. arXiv preprint arXiv:2206.07682  (2022)

\bibitem{VisualChatGPT}
Wu, C., Yin, S., Qi, W., Wang, X., Tang, Z., Duan, N.: Visual chatgpt: Talking,
  drawing and editing with visual foundation models. arXiv preprint
  arXiv:2303.04671  (2023)

\bibitem{yang2023gpt4tools}
Yang, R., Song, L., Li, Y., Zhao, S., Ge, Y., Li, X., Shan, Y.: Gpt4tools:
  Teaching large language model to use tools via self-instruction. arXiv
  preprint arXiv:2305.18752  (2023)

\bibitem{MMREACT}
Yang, Z., Li, L., Wang, J., Lin, K., Azarnasab, E., Ahmed, F., Liu, Z., Liu,
  C., Zeng, M., Wang, L.: Mm-react: Prompting chatgpt for multimodal reasoning
  and action. arXiv preprint arXiv:2303.11381  (2023)

\bibitem{GLM}
Zeng, A., Liu, X., Du, Z., Wang, Z., Lai, H., Ding, M., Yang, Z., Xu, Y.,
  Zheng, W., Xia, X., et~al.: Glm-130b: An open bilingual pre-trained model.
  arXiv preprint arXiv:2210.02414  (2022)

\bibitem{zhang2022dino}
Zhang, H., Li, F., Liu, S., Zhang, L., Su, H., Zhu, J., Ni, L.M., Shum, H.Y.:
  Dino: Detr with improved denoising anchor boxes for end-to-end object
  detection. arXiv preprint arXiv:2203.03605  (2022)

\bibitem{OPT}
Zhang, S., Roller, S., Goyal, N., Artetxe, M., Chen, M., Chen, S., Dewan, C.,
  Diab, M., Li, X., Lin, X.V., et~al.: Opt: Open pre-trained transformer
  language models. arXiv preprint arXiv:2205.01068  (2022)

\bibitem{minigpt4}
Zhu, D., Chen, J., Shen, X., Li, X., Elhoseiny, M.: Minigpt-4: Enhancing
  vision-language understanding with advanced large language models. arXiv
  preprint arXiv:2304.10592  (2023)

\end{thebibliography}
}
\newpage

\appendix

\section{Appendix}

\subsection{More Interactive Cases}
\label{more cases}

\begin{figure*}[hbtp]
\centering
\vspace{8pt}
\includegraphics[width=1.0\linewidth]{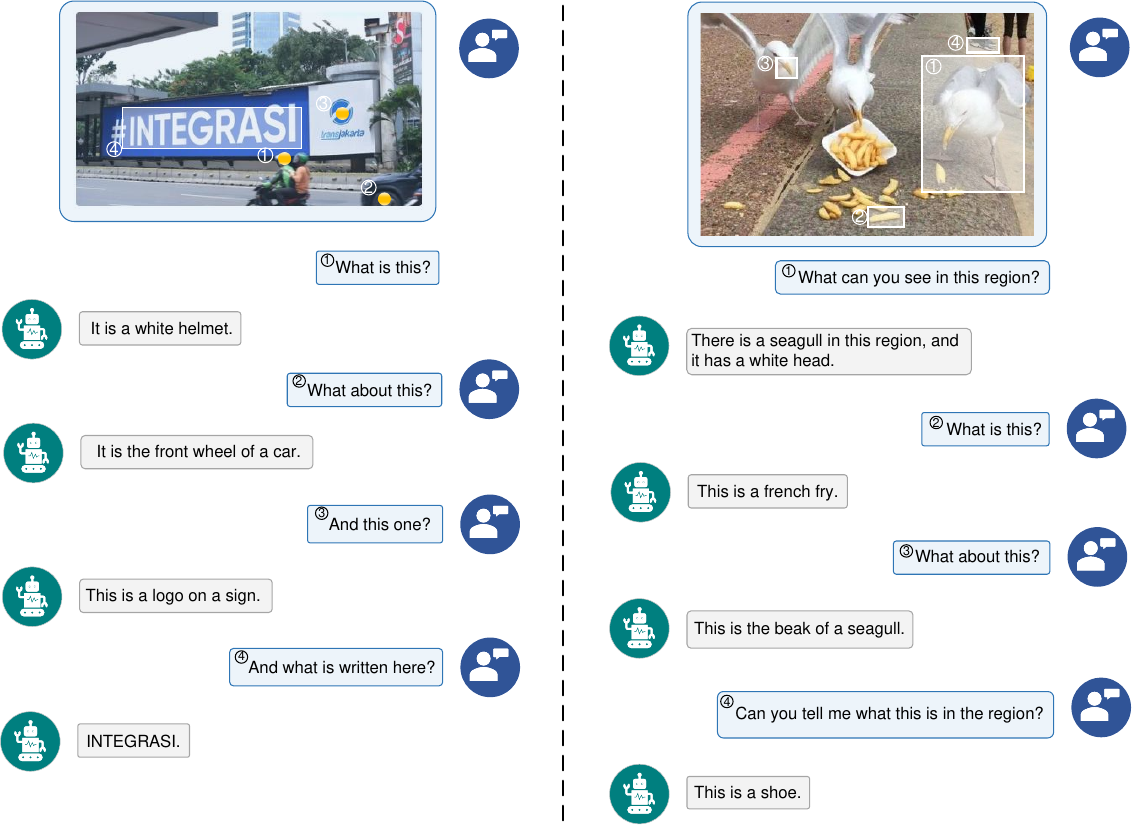}
\caption{More cases of interacting with ChatSpot.}
\label{fig.8}
\end{figure*}

\begin{figure*}[hbtp]
\centering
\includegraphics[width=1.0\linewidth]{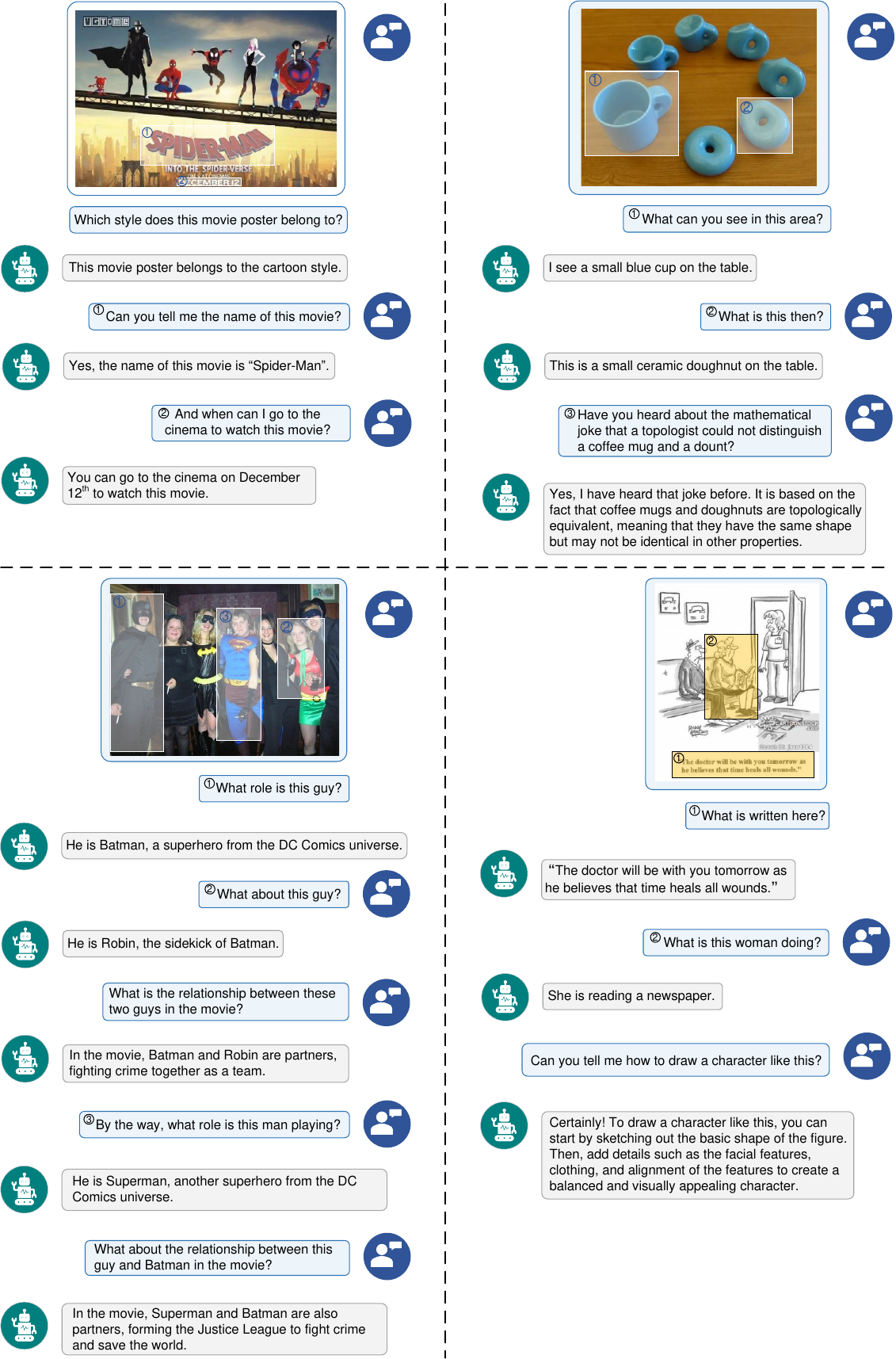}
\caption{More cases of interacting with ChatSpot.}
\label{fig.7}
\end{figure*}

In this section, we provide additional dialogue records of ChatSpot in this section. As shown in Figure \ref{fig.8} and Figure \ref{fig.7}. ChatSpot supports multiple levels of interaction, including full image, region boxes, and region points. In the future, we will support more diverse interaction forms.

\subsection{Failure Cases Analysis}

\begin{figure*}[hbtp]
\centering
\includegraphics[width=1.0\linewidth]{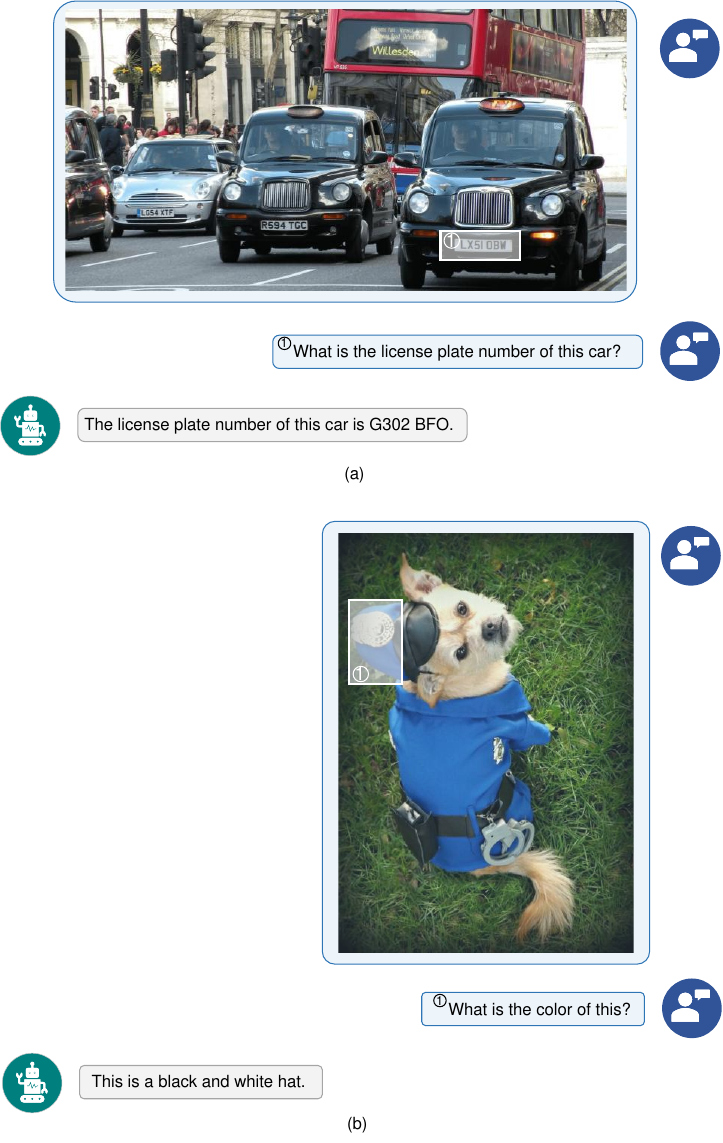}
\caption{Some failure cases of ChatSpot. (a) ChatSpot on license plate recognition. (b) ChatSpot on color recognition.}
\label{fig.9}
\end{figure*}

Due to the limitation of data and instructions, ChatSpot may encounter challenges in certain landmark scenarios. We have conducted a comprehensive analysis and are determined to address these limitations in the next version. Specifically, as shown in Figure \ref{fig.9} (a), due to the limited inclusion of license plate recognition data in our dataset MGVLID, ChatSpot's performance in license plate recognition is not satisfactory. And, as illustrated in Figure \ref{fig.9} (b), ChatSpot may encounter challenges in accurately identifying the precise colors of regions or objects when confronted with complex backgrounds. To address these failure cases, we can overcome them by incorporating specific instructions. In future versions, we aim to develop a wider range of instructions while also ensuring a balanced dataset.

\end{document}